\documentclass[preprint,12pt,authoryear]{elsarticle}

\usepackage{amssymb}
\usepackage{amsmath}

\usepackage[noend]{algpseudocode}
\usepackage{algorithm}
\usepackage{multirow}
\usepackage{color,xcolor}
\usepackage{comment}
\usepackage{bbding}
\journal{***}

\begin{document}

\begin{frontmatter}



\title{One Model for Two Tasks: Cooperatively Recognizing and Recovering Low-Resolution Scene Text Images by Iterative Mutual Guidance} 



\author[fudan]{Minyi Zhao}
\ead{zhaomy20@fudan.edu.cn}
\author[tongji]{Yang Wang}
\ead{tongji\_wangyang@tongji.edu.cn}
\author[tongji]{Jihong Guan}
\ead{jhguan@tongji.edu.cn}
\author[fudan]{Shuigeng Zhou\corref{mycorrespondingauthor}}
\cortext[mycorrespondingauthor]{Corresponding author}
\ead{sgzhou@fudan.edu.cn}
\affiliation[fudan]{organization={Shanghai Key Lab of Intelligent Information Processing, and School of Computer Science, Fudan University},
            country={China}}
\affiliation[tongji]{organization={Department of Computer Science and Technology, Tongji University},
            country={China}}

\begin{abstract}
Scene text recognition (STR) from high-resolution (HR) images has been significantly successful, however text reading on low-resolution (LR) images is still challenging due to  insufficient visual information. Therefore, recently many scene text image super-resolution (STISR) models have been proposed to generate super-resolution (SR) images for the LR ones, then STR is done on the SR images, which thus boosts recognition performance. Nevertheless, these methods have two major weaknesses. On the one hand, STISR approaches may generate imperfect or even erroneous SR images, which mislead the subsequent recognition of STR models. On the other hand, as the STISR and STR models are jointly optimized, to pursue high recognition accuracy, the fidelity of SR images may be spoiled. As a result, neither the recognition performance nor the fidelity of STISR models are desirable. Then, can we achieve both high recognition performance and good fidelity? To this end,
in this paper we propose a novel method called \textbf{IMAGE} (the abbreviation of \textbf{I}terative \textbf{M}utu\textbf{A}l \textbf{G}uidanc\textbf{E}) to effectively recognize and recover LR scene text images simultaneously. Concretely, IMAGE consists of a specialized STR model for recognition and a tailored STISR model to recover LR images, which are optimized separately. And we develop an iterative mutual guidance mechanism, with which the STR model provides high-level semantic information as clue to the STISR model for better super-resolution, meanwhile the STISR model offers essential low-level pixel clue to the STR model for more accurate recognition. Extensive experiments on two LR datasets demonstrate the superiority of our method over the existing works on both recognition performance and super-resolution fidelity.
\end{abstract}




\end{frontmatter}


\section{Introduction}
Texts in images offer essential information, which can be extracted and interpreted for many text-based understanding tasks (\textit{e.g.,} text-VQA~\cite{singh2019towards}, Doc-VQA~\cite{mathew2021docvqa}, and ViteVQA~\cite{zhaotowards}) and various industrial scenarios  (\textit{e.g.,} tag recognition~\cite{deng2023end}, license-plate recognition~\cite{kong2021federated}, and vehicle identification~\cite{yin2023mirrored}). Although scene text recognition (STR) from high-resolution (HR) images has achieved great progress and successfully applied to various scenarios (e.g. car plate recognition and bank card recognition) due to a series of advanced scene text recognizers~\cite{du2022svtr} developed via deep learning in recent years, the performance of reading low-resolution (LR) scene text images  is still unsatisfactory, due to various image quality degradation like blurry and low-contrast~\cite{wang2020scene,ma2023benchmark}.

\begin{figure*}
	\begin{center}
		\includegraphics[width=0.65\linewidth]{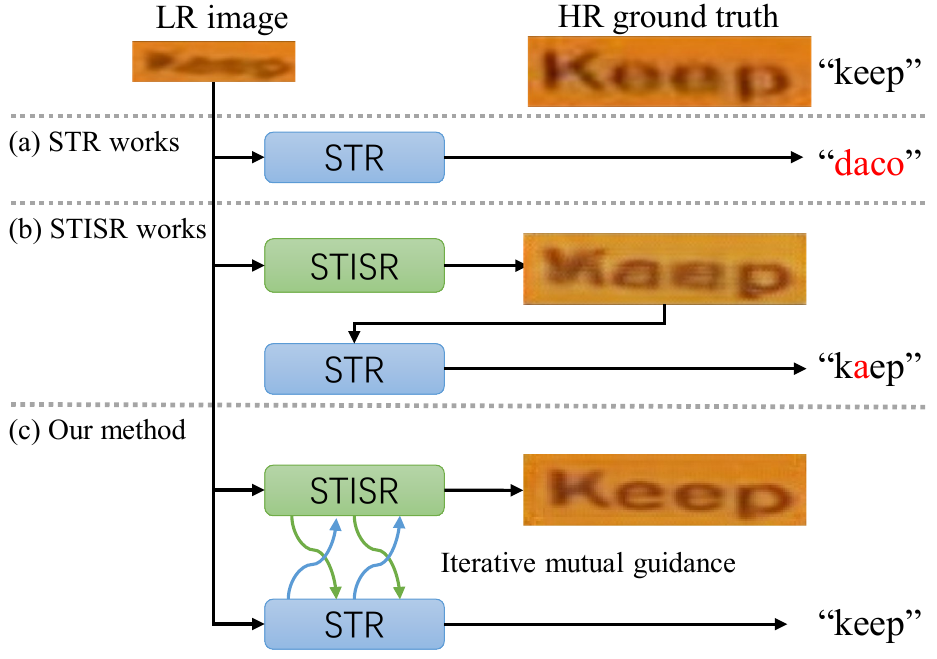}
	\end{center}
	\caption{Schematic illustration of existing works: (a) STR methods and (b) STISR methods,
and (c) our method IMAGE that employs two models to do recognition and recovery respectively, which are optimized separately, but mutually provide guidance clues to each other in an iterative way. The rightmost character strings are the recognition results, where red characters are incorrectly recognized, and black ones are correctly recognized.}
	\label{fig:motivation}
\end{figure*}

As shown in Fig.~\ref{fig:motivation}, existing approaches to read LR text images can be categorized into two groups. The first group directly recognizes texts from LR images, most early STR works fall into this group. However, as pointed out in \cite{wang2020scene,zhao2022c3}, due to the lack of sufficient pixel information, they are prone to misread the texts. As a result, the text is incorrectly recognized (see Fig.~\ref{fig:motivation}(a)). Another group tries to design a scene text image super-resolution (STISR)~\cite{lan2020cascading,lan2020madnet,jiang2019ensemble,chen2022text,wang2020scene} model to serve as a pre-processing technique to recover the missing details in LR images for boosting text recognition performance as well as the visual quality of the scene texts. More and more recently proposed models follow this line. However, these STISR methods may generate imperfect or even erroneous SR images, which will mislead the subsequent recognition. Besides, in STISR works,
as the STISR and STR models are jointly optimized, for pursuing high recognition accuracy, the fidelity of SR images may be degraded. For example, as shown in Fig.~\ref{fig:motivation}(b), not only the first `e' is incorrectly recovered, which misleads the subsequent recognition, but also the contrast is enhanced to highlight the texts, which impairs the fidelity of the SR image. As a result, both the recognition performance and the fidelity of STISR models are hurt.

Then, \emph{can we solve the aforementioned problems and achieve win-win in both recognition performance and super-resolution fidelity}? For this purpose, in this paper we propose a new method called \textbf{IMAGE} (the abbreviation of \textbf{I}terative \textbf{M}utu\textbf{A}l \textbf{G}uidanc\textbf{E}) to effectively recognize and recover LR scene text images simultaneously, as shown in Fig.~\ref{fig:motivation}(c). IMAGE consists of two models: a recognition model to read LR images and a super-resolution model aims at recovering the LR images. The two models are optimized separately to avoid malign competition between them. Meanwhile, we develop an iterative mutual guidance mechanism, with which the STR model provides high-level semantic information as clue to the STISR model for better super-resolution, meanwhile the STISR model offers essential low-level pixel clue to the STR model for more accurate recognition. This mechanism is illustrated by the blue and green curves in Fig.~\ref{fig:motivation}(c). The iterative interaction between the recognition model and the STISR model helps them reach their own goals cooperatively.

Contributions of this paper are as follows:
\begin{itemize}
    \item Observing the limitations of existing STR and STISR models, we propose a new solution to recognize and recover LR text images simultaneously.
    \item We develop an iterative mutual guidance mechanism to coordinate the two models so that they are optimized separately, while help each other to reach their own goals by providing informative clues.
    \item We conduct extensive experiments on two LR datasets, which show the advantages of IMAGE over existing techniques on both recognition performance and super-resolution fidelity.
\end{itemize}

\section{Related work}
Here we briefly review the super-resolution techniques and some typical scene text recognizers, and highlight the differences between existing works and our method.

\subsection{Scene text image super-resolution}
According to whether exploiting text-specific information from HR images, recent STISR methods can be roughly divided into two types: generic super-resolution approaches~\cite{ma2017learning,ates2023deep} and scene text image super-resolution approaches. Generic image super-resolution methods usually use pixel information captured by pixel loss functions (\textit{i.e.,} L1 or L2 loss) to supervise their models. In particular, SRCNN~\cite{dong2015image,pandey2018binary} designs various convolutional neural networks. \cite{xu2017learning} and SRResNet~\cite{ledig2017photo} adopt generative adversarial networks.  RCAN~\cite{zhang2018image} and SAN~\cite{dai2019second} introduce attention mechanisms to advance the recovery. Recently, transformer-structured methods~\cite{li2021efficient,liang2021swinir,wang2022uformer} are proposed to further boost performance. Differently, STISR approaches focus on extracting various text-specific information to help the model. Specifically, \cite{fang2021tsrgan,wang2019textsr,qin2022scene} calculate text-specific losses. \cite{mou2020plugnet} introduces a pluggable super-resolution module. \cite{wang2020scene} employs TSRN and gradient profile loss to capture sequential information of text images and gradient
fields of HR images for sharpening the texts. PCAN~\cite{zhao2021scene} proposes to learn sequence-dependent and high-frequency information for recovery. STT~\cite{chen2021scene} uses a pre-trained transformer recognizer for text-focused super-resolution. TG~\cite{chen2022text} exploits stroke-level information from HR images for more fine-grained super-resolution. TPGSR~\cite{ma2023text}, TATT~\cite{ma2022text}, C3-STISR~\cite{zhao2022c3}, and \cite{huang2023text} make use of various text-specific clues to guide the super-resolution. Nevertheless, these STISR methods have the drawbacks of forcing STR models to read erroneous SR images and not being able to balance recognition performance and fidelity on SR images. Thereby, both the recognition ability and fidelity are unsatisfactory.

\subsection{Scene text recognition}
Text recognition on HR images has made great progress in recent years~\cite{bai2018edit,chen2021benchmarking,chen2021survey,cheng2017focusing,cheng2018aon}. Specifically, CRNN~\cite{shi2016end} takes CNN and RNN as the encoder and proposes a CTC-based~\cite{graves2006connectionist} decoder. ASTER~\cite{shi2018aster} introduces a spatial transformer network (STN)~\cite{jaderberg2015spatial} to rectify irregular text images. MORAN~\cite{luo2019moran} proposes a multi-object rectification network. \cite{hu2020gtc,sheng2019nrtr,yu2020towards} use different attention mechanisms. AutoSTR~\cite{zhang2020efficient} employs neural architecture search (NAS)~\cite{elsken2019neural} to search backbone. More recently, semantic-based~\cite{qiao2020seed,yu2020towards}, transformer-based~\cite{atienza2021vision}, linguistical-based~\cite{fang2021read,wang2021two}, and high-efficiency~\cite{du2022svtr,bautista2022parseq} approaches are proposed to further lift the performance. Although these methods are able to handle irregular, occluded, and incomplete text images, they still have difficulty in recognizing low-resolution images due to the lack of sufficient pixel information. What is more, finetuning these recognizers is insufficient to accurately recognize texts from LR images, as reported in~\cite{wang2020scene}. Therefore, existing STR techniques can not handle text reading of LR images well.
\begin{figure*}
	\begin{center}
		\includegraphics[width=0.98\linewidth]{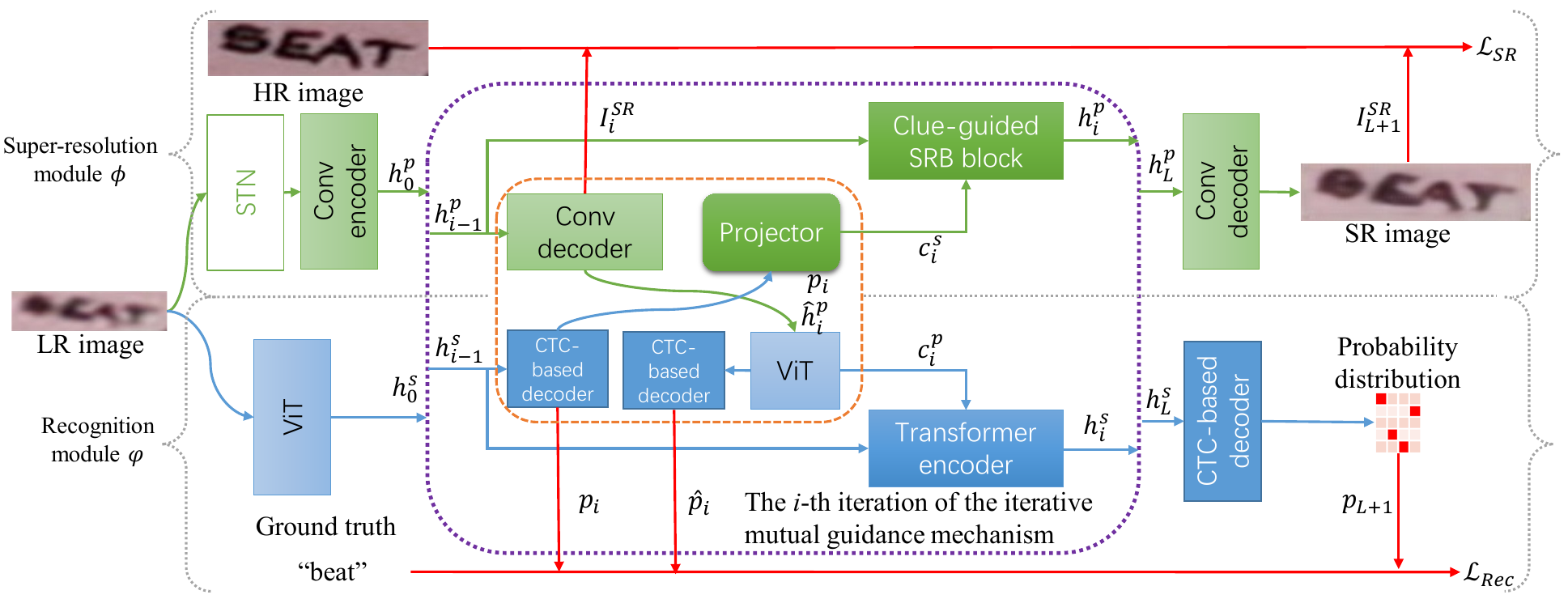}
	\end{center}
	\caption{The architecture of our method IMAGE.}
	\label{fig:pipeline}
\end{figure*}
\subsection{Differences between related works and our method}
First, comparing with recent STR methods, IMAGE not only takes the advantage of pixel information to aid recognition but also outputs an SR image of high fidelity. Second, in contrast to STISR models that require STR models to directly read the SR images that may carry noisy or erroneous information, we provide low-level information as indirect clue to aid the recognition. Besides, we relax the requirement of lifting recognition performance on the STISR model so that it can focus on improving fidelity. Our iterative mutual guidance mechanism lets the STISR and STR models help each other cooperatively. As a result, IMAGE outperforms STISR models on both recognition and fidelity.

\section{Methodology}
Here we first give an overview of our method IMAGE, then present the design of super-resolution model and the text recognition model. Subsequently, we introduce the iterative mutual guidance mechanism in detail, followed by the design of loss function and the overall training procedure to better demonstrate the implementation.
\subsection{Overview}
\label{sec:overview}
Given a low-resolution image $I_{LR}$ $\in$ $\mathbb{R}^{C \times N}$. Here, $C$ is the number of channels of each image, $N$ = $H \times W$ is the
collapsed spatial dimension, $H$ and $W$ are the height and width of the LR image. Our aim is to produce a super-resolution (SR) image $I_{SR}$ $\in$ $\mathbb{R}^{C \times (4 \times N)}$ based on the input LR image $I_{LR}$ and recognize the text on $I_{LR}$.

Fig.~\ref{fig:pipeline} shows the architecture of our method IMAGE, which is composed of three major components: the \emph{super-resolution model} $\phi$ (painted in green) aims at recovering the LR image, the \emph{recognition model} $\varphi$ (colored in blue) recognizes the text from the LR image, and an \emph{iterative mutual guidance mechanism} (colored in orange), which is the core component of IMAGE, is used to bridge $\phi$ and $\varphi$ and extract useful clues to guide the recognition and recovery. To achieve a more comprehensive interaction, such mechanism is conducted repeatedly over $L$ iterations.

In the $i$-th iteration, two features from the last iteration are taken as input, namely \emph{pixel feature} $h_{i-1}^{p} \in \mathbb{R}^{C' \times N}$ where $C'$ is the intermediate channel number, and \emph{semantic feature} $h_{i-1}^{s} \in \mathbb{R}^{M \times D}$ where $M$ and $D$ denote the number of the tokens and the dimensionality of the tokens, respectively, to produce two clues to aid the corresponding tasks, denoted as \emph{semantic clue} $c_i^s \in \mathbb{R}^{C' \times N}$ for super-resolution and \emph{pixel clue} $c_i^p \in \mathbb{R}^{M \times D}$ for text recognition.

Furthermore, we generate three intermediate results, including an SR image $I^{SR}_{i}$, a predicted probability distribution $p_{i} \in \mathbb{R}^{M \times |\mathcal{A}|}$ of the last semantic feature $h_{i-1}^s$ where $|\mathcal{A}|$ is the size of a charset $\mathcal{A}$, and a probability distribution $\hat{p}_{i} \in \mathbb{R}^{M \times |\mathcal{A}|}$ of the pixel clue $c_i^p$, for the supervision of model training. After generating clues, we combine the clues with the last features (\textit{i.e.,} $h_{i-1}^p$ for $\phi$ and $h_{i-1}^s$ for $\varphi$) to generate the features $h_{i}^p$ and $h_{i}^s$. Eventually, two decoders are stacked to decode the final pixel feature $h_{L}^p$ and semantic feature $h_{L}^s$ for the generation of the final SR image $I_{L+1}^{SR}$ and the prediction $p_{L+1}$.

In IMAGE, a total of $L+1$ SR images, denoted as $\{I^{SR}_{1}, ..., I^{SR}_{L+1}\}$, and $2L+1$ probability distributions (\textit{i.e.,} $\{p_{1}, ..., p_{L+1}\}$ and $\{\hat{p}_{1}, ..., \hat{p}_{L}\}$) are generated. We use the HR image $I_{HR}$ of each training LR image and its text-level annotation $p_{GT}$, all the generated SR images and predicted probability distributions to evaluate the super-resolution loss $\mathcal{L}_{SR}$ and the recognition loss $\mathcal{L}_{Rec}$ for model training.

\subsection{Super-resolution model}
\label{sec:sr}
Different from the existing STISR models, a semantic clue is additionally input to our super-resolution model $\phi$ to aid the recovery. Thereby, we design the super-resolution model $\phi$ in the following way: 1) Begin with a Spatial Transformer Network (STN) to address the pixel-level offsets caused by the manual alignment of LR-HR image pairs~\cite{wang2020scene,chen2021scene,chen2022text,ma2023text}. 2) A convolution encoder is attached to provide the initial pixel feature $h_0^p$ for super-resolution. 3) $L$ clue-guided SRB blocks~\cite{ma2023text,zhao2022c3}, each of which consists of convolution layers and two GRUs to capture sequential information, are subsequently stacked. For the $i$-th SRB block, we use the last iteration's feature $h_{i-1}^p$ and the semantic clue $c_{i}^s$ to conduct a semantic clue guided super-resolution. Formally, let the $i$-th SRB block be $f_i$, we have
\begin{equation}
\label{eq:srb}
h_{i}^{p} = f_{i}(h_{i-1}^p,c_{i}^s).
\end{equation}
Following \cite{ma2023text,zhao2022c3}, the clue is concatenated with the feature map extracted by the convolution layers of the SRB block at channel dimension. 4) A convolution decoder made up of several convolution layers and a pixel shuffle module is attached at the end of $L$ clue-guided SRB blocks to form the final SR Image $I_{L+1}^{SR}$.

\subsection{Text recognition model}
\label{sec:rec}
In this section, we introduce the design of the text recognition model $\varphi$ used in IMAGE. Our recognition model is also different from typical STR models to support the recognition of LR images and is additionally fed in a pixel clue to boost the recognition. Notice that the major challenge of reading LR image texts is the lack of sufficient pixel information. Accordingly, we use transformer-based structure, instead of some convolution and pooling operations used in \cite{shi2016end,du2022svtr}, to implement the recognizer, which performs better at capturing limited and valuable pixel information thanks to the multi-head self-attention mechanism. In particular, similar to the super-resolution model, we first use a ViT with a patch size of $[H,W//32]$ to offer an initial semantic feature $h_0^s$. In what follows, $L$ transformer encoders are stacked to fuse the recognition feature and the pixel clue. Taking the $i$-th iteration for instance, let the ($i$-1)-th semantic feature, current pixel clue, and the $i$-th transformer encoder layers be $h_{i-1}^s$ and $c_i^p$, and $g_i$, respectively. We compute the $i$-th recognition feature as follows:
\begin{equation}
\label{eq:trans}
h_{i}^{s} = g_{i}([h_{i-1}^s,c_{i}^p]),
\end{equation}
where $[\cdot, \cdot]$ denotes the concatenating operation. Finally, a CTC-based decoder is employed to predict the final probability distribution $p_{L+1}$. After decoding $p_{L+1}$ on the charset $\mathcal{A}$, the text is obtained.

\subsection{Iterative mutual guidance}
\label{sec:img}
We have introduced how the super-resolution model $\phi$ and the recognition model $\varphi$ are designed to support SR images generation and LR images recognition. In this section, we describe how the iterative mutual guidance mechanism provides the clues to $\phi$ and $\varphi$. In particular, taking the $i$-th iteration for example, we denote the $i$-th mutual guidance as $u_i$, which generates the semantic clue $c_{i}^s$ and the pixel clue $c_{i}^p$ according to two previous features $h_{i-1}^{p}$ and $h_{i-1}^{s}$, and additionally outputs three intermediate results, including an SR image $I_{i}^{SR}$ and two probability distributions $p_{i}$ and $\hat{p}_{i}$, for better representations of the clues. Formally, we have
\begin{equation}
\label{eq:cgm}
c_{i}^s, c_{i}^p, I_{i}^{SR}, p_{i}, \hat{p}_{i}  = u_{i}(h_{i-1}^p,h_{i-1}^s).
\end{equation}
Notice that there is a modality gap between $h_{i-1}^{p}$ and $h_{i-1}^{s}$ ($H\times W$ for the pixel feature while $M \times |\mathcal{A}|$ for the semantic feature). Hence, some cross-modal processing techniques should be used to convert the modalities.
Specifically, as shown in the orange rectangle in Fig.~\ref{fig:pipeline}, five networks are used: a convolution decoder $q^{CD}_{i}:=\mathbb{R}^{C' \times N} \rightarrow \mathbb{R}^{C \times N}$, a projector $q^{OJ}_{i}:=\mathbb{R}^{M \times |\mathcal{A}|} \rightarrow \mathbb{R}^{C' \times N}$, two CTC-based decoders $q^{CT1}_{i}$ and $q^{CT2}_{i}$ ($q^{CT1}_{i}:=\mathbb{R}^{M \times D} \rightarrow \mathbb{R}^{M \times |\mathcal{A}|}$, $q^{CT2}_{i}:=\mathbb{R}^{M \times D} \rightarrow \mathbb{R}^{M \times |\mathcal{A}|}$), and a ViT $q^{VT}_{i}:=\mathbb{R}^{C' \times N} \rightarrow \mathbb{R}^{M \times D}$.

For the generation of semantic clue $c_{i}^s$ to guide super-resolution model, we first use a CTC-based decoder implemented by a fully connection layer to extract the predicted probability distribution:
\begin{equation}
\label{eq:ct}
p_{i} = q^{CT1}_{i}(h_{i-1}^s).
\end{equation}
Then, a projector $q^{PJ}_{i}$ made up of four
transposed convolution layers followed by batch normalization and a bilinear interpolation is used to project the high-level feature into pixel-level:
\begin{equation}
\label{eq:pj}
c_{i}^{s} = q^{PJ}_{i}(p_{i}).
\end{equation}

The processing of the pixel feature $h_{i-1}^{p}$ is performed as follows: First, a convolution decoder $q^{CD}_{i}$ is used to generate the SR image according to the feature $h_{i-1}^{p}$:
\begin{equation}
\label{eq:cd}
\hat{h}_{i}^s, I_{i}^{SR} = q^{CD}_{i}(h_{i-1}^{p}),
\end{equation}
where $\hat{h}_{i}^s \in \mathbb{R}^{C' \times N}$ is the feature output by the pixel shuffle module in the convolution decoder $q^{CD}_{i}$. Then, a ViT with a patch size of $[2*H,W//16]$ is used to generate the clue containing richer pixel information for the recognition model:
\begin{equation}
\label{eq:vt}
c_{i}^p = q^{VT}_i(\hat{h}_{i}^s).
\end{equation}
Finally, to make the pixel clue much more representative, we explicitly use a CTC-based decoder $q^{CT2}_{i}$ to offer an intermediate result $\hat{p}_{i}$ for supervision:
\begin{equation}
\label{eq:ct2}
\hat{p}_{i} = q^{CT2}_{i}(c_{i}^p).
\end{equation}

\subsection{Loss functions}
Two losses are computed to supervise the model training. It is worth mentioning that we calculate the losses for all the intermediate results generated in Sec.~\ref{sec:img} to provide the supervision for the representations of the two clues. The first is super-resolution loss $\mathcal{L}_{SR}$ used for $\phi$, which can be calculated via the gradient profile loss $\mathcal{L}_{GP}$~\cite{wang2020scene} between HR image $I_{HR}$ and all SR images:
\begin{equation}
\label{eq:sr_loss}
\mathcal{L}_{SR} = \Sigma_{i}2^{i}\mathcal{L}_{GP}(I_{i}^{SR},I_{HR}).
\end{equation}
A weight $2^{i}$ is used to make the model to put more emphasis on deeper iterations. Besides, a recognition loss $\mathcal{L}_{Rec}$ is calculated to supervise the recognition module by the CTC loss $\mathcal{L}_{CTC}$ between the ground truth $p_{GT}$ and all the intermediate probability distributions:
\begin{equation}
\label{eq:rec_loss}
\mathcal{L}_{Rec} = \Sigma_{i}2^{i}\mathcal{L}_{CTC}(p_{i},p_{GT}) + \Sigma_{i}2^{i}\mathcal{L}_{CTC}(\hat{p}_{i},p_{GT}).
\end{equation}
The final loss of IMAGE is the combination of the super-resolution loss $\mathcal{L}_{SR}$ and the recognition loss $\mathcal{L}_{Rec}$:
\begin{equation}
\label{eq:final_loss}
\mathcal{L} = \mathcal{L}_{SR} + \mathcal{L}_{Rec}.
\end{equation}
\begin{algorithm}[t]
\caption{The training procedure of IMAGE.}
\begin{algorithmic}[1]
\State {\bf Input:}
$\phi, \varphi, I_{LR}, I_{HR}, p_{GT}, L$
\State Generate initial $h_0^p$ and $h_0^s$ as described in Sec.~\ref{sec:sr} and Sec.~\ref{sec:rec}, respectively
\For{$i$ in $[1,...,L]$}
    \State $c_{i}^s, c_{i}^p, I_{i}^{SR}, p_{i},  \hat{p}_{i}  = u_{i}(h_{i-1}^p,h_{i-1}^s)$
    \State $h_{i}^{p} = f_{i}(h_{i-1}^p,c_{i}^s)$
    \State $h_{i}^{s} = g_{i}([h_{i-1}^s,c_{i}^p])$
\EndFor
\State Generate the final output $I_{L+1}^{SR}$ and $p_{L+1}$ according to Sec.~\ref{sec:sr} and Sec.~\ref{sec:rec}, respectively
\State Compute the loss $\mathcal{L}$ via Eq.~(\ref{eq:final_loss})
\State \Return $\mathcal{L}$
\end{algorithmic}
\label{alg:image}
\end{algorithm}

\subsection{Overall training procedure}
Here we describe the overall training procedure of IMAGE to better demonstrate the implementation. As presented in Alg.~\ref{alg:image}, we first generate the initial pixel feature $h_0^p$ and the initial semantic feature $h_0^s$. Then, $L$ rounds of iterative mutual guidance (Line 3$ - $Line 6) introduced in Sec.~\ref{sec:img} are performed to conduct a clue-guided super-resolution and recognition. Subsequently, the last SR image $I_{L+1}^{SR}$ and predicted probability distribution $p_{L+1}$ are generated (Line 7). Finally, the loss of IMAGE is computed to supervise the model training (Line 8).

\section{Performance Evaluation}
In this section, we first introduce the datasets and metrics used in the experiments and the implementation details. Then, we evaluate IMAGE and compare it with state-of-the-art techniques to show its effectiveness and superiority. Finally, we conduct extensive ablation studies to validate the design of our method.

\subsection{Datasets and metrics}
Following recent STR and STISR works~\cite{shi2016end,chen2021scene,ma2022text,zhao2022c3}, we use the synthetic datasets to pre-train our model and then finetune the model on the low-resolution datasets. The synthetic datasets include Synth90K~\cite{jaderberg2014synthetic} and SynthText~\cite{gupta2016synthetic}.
Two real-world low-resolution datasets are TextZoom and IC15-352. In what follows, we give a brief introduction to the low-resolution datasets used for finetuning and performance evaluation.

The \textbf{TextZoom}~\cite{wang2020scene} dataset consists of 21,740 LR-HR text image pairs collected by lens zooming of the camera in real-world scenarios. The training set has 17,367 pairs, while the test set is divided into three settings based on the camera focal length: easy (1,619 samples), medium (1,411 samples), and hard (1,343 samples).

The \textbf{IC15-352}~\cite{chen2022text} dataset consists of 352 low-resolution images. This dataset is used to check the generalization power of our model trained on TextZoom when being adapted to other datasets.

There are six metrics used in our paper to evaluate the performance of the model from the aspects of recognition ability, fidelity, and efficiency, accordingly. The first metric is \emph{word-level recognition accuracy} which evaluates the recognition performance of various methods. Following the settings of previous
works~\cite{chen2022text,ma2023text,ma2022text,zhao2022c3}, we remove punctuation and convert uppercase letters to lowercase letters for calculating recognition accuracy. Besides, we report \emph{Peak Signal-to-Noise Ratio} (PSNR) and \emph{Structure Similarity Index Measure} (SSIM)~\cite{wang2004image} as the metrics to measure the fidelity performance. As for efficiency, we calculate the \textit{Floating-point Operations per Second} (FLOPs), the number of parameters (Params) of each model, and the inference time cost (ms) to measure the computational cost.

\subsection{Implementation details}
All experiments are conducted on 4 NVIDIA RTX 3090 GPUs with 24GB memory. The PyTorch version is 1.10. IMAGE is trained using AdamW optimizer with a learning rate of 0.0001. The batch
size is set to 256 for pre-training and 64 for finetuning. The intermediate channel number $C'$ is set to 64. $M$ and $D$ are set to $32$ and $196$, respectively. Following \cite{du2022svtr,shi2016end,zhao2022c3,chen2022text}, we use a charset with a length of 36 (\textit{i.e.,} letters and numbers). To better demonstrate the advantage of our technique over different computational requirements, IMAGE with various iteration numbers are evaluated, denoted as IMAGE-$L$, \textit{e.g.,} for $L=2$ we have IMAGE-2.
\begin{table*}[t]
\centering
\newsavebox{\overalll}
\begin{lrbox}{\overalll}
\begin{tabular}{c|c|c|cccc||c|c||c|c}
\hline
\multirow{2}*{ID}&\multirow{2}*{Method}&\multirow{2}*{Datasets}& \multicolumn{4}{c||}{Recognition accuracy}
& \multicolumn{2}{c||}{Fidelity performance}& \multicolumn{2}{c}{Computational cost}\\
\cline{4-11}
~&~&~& Easy& Medium & Hard&Average &PSNR~(dB)&SSIM~($\times10^{-2}$)&FLOPs~(G)&Params~(M)
\\
\hline
1&LR&- &- &- &- &- &20.35 &69.61 &- &-\\
\hline
\hline
2&CRNN&S &37.5\% &21.4\% &21.1\% &27.3\% &- &- &0.91 &8.36\\
3&SVTR&S &69.1\% &45.6\% &34.1\% &50.8\% &- &- &1.90 &22.70\\
\hline
4&CRNN&S+L  &58.4\% &34.1\% &27.6\% &41.1\% &- &- &0.91 &8.36\\
5&SVTR&S+L &80.0\% &59.0\% &42.4\% &61.7\% &- &- &1.90 &22.70\\
\hline
\hline
6&TG$^{1}$&S+L &61.2\% &47.6\% &35.5\% &48.9\% &21.40 &74.56 &1.82 &9.19\\
7&TPGSR$^{1}$&S+L &63.1\% &52.0\% &38.6\% &51.8\% &21.18 &\textbf{77.62} &4.75 &35.33\\
8&C3-STISR$^{1}$&S+L &65.2\% &53.6\% &39.8\% &53.7\% &21.51 &77.21 &4.54 &58.52\\
\hline
9&TG$^{2}$&S+L &76.8\% &61.5\% &43.3\% &61.6\% &21.40 &74.56 &2.81 &23.53\\
10&TPGSR$^{2}$&S+L &80.9\% &62.8\% &46.4\% &64.5\% &21.18 &\textbf{77.62} &5.74 &49.67\\
11&C3-STISR$^{2}$&S+L &81.6\% &63.8\% &46.7\% &65.1\% &21.51 &77.21 &5.53 &72.86\\
\hline
\hline
12&IMAGE-1&S &72.5\% &56.1\% &41.0\% &57.5\% &19.52 &62.47 &1.98 &31.84\\
13&IMAGE-1&S+L &\textbf{81.8}\% &\textbf{68.9}\% &\textbf{52.2}\% &\textbf{68.6}\% &21.17 &71.16 &1.98 &31.84\\
\hline
14&IMAGE-2&S &72.7\% &59.0\% &41.6\% &58.7\% &19.75 &63.16 &3.41 &56.22\\
15&IMAGE-2&S+L &\textbf{83.6}\% &\textbf{70.0}\% &\textbf{54.8}\% &\textbf{70.3}\% &\textbf{21.92} &74.56 &3.41 &56.22\\
\hline
\end{tabular}
\end{lrbox}
\scalebox{0.55}{\usebox{\overalll}}
\caption{Performance comparison with various methods on the TextZoom dataset. ``Datasets'' denotes the datasets used to develop the model. Concretely, `S': synthetic dataset; `L': low-resolution dataset. $^1$: CRNN as recognizer. $^2$: SVTR as recognizer.}
\label{tab:textzoom}
\end{table*}

\subsection{Comparing with SOTA approaches}
\subsubsection{Performance improvement on TextZoom}
Here, we evaluate our method on the \textbf{TextZoom} dataset. In particular, we compare our method with two recognizers including one typical method widely used by recent STISR methods: CRNN~\cite{shi2016end} and one recently proposed state-of-the-art recognizer SVTR~\cite{du2022svtr} and three recent STISR methods including TG~\cite{chen2022text}, TPGSR~\cite{ma2023text}, and C3-STISR~\cite{zhao2022c3}. All the results are presented in Tab.~\ref{tab:textzoom}.

We start by analyzing the performance of existing STR and STISR methods. From Tab.~\ref{tab:textzoom}, we can see that (1) Typical STR models cannot do the task of low-resolution scene text image recognition well. As can be checked in the 2nd row and the 3rd row, CRNN and the state-of-the-art recognizer SVTR can only correctly read 27.3\% and 50.8\% images, respectively. In the rest of this paper, we only discuss two typical and efficient recognizers (\textit{i.e.,} CRNN and SVTR).
(2) Finetuning these recognizers still cannot do the work well. For example, after finetuning, the recognition performance of CRNN and SVTR is boosted from 27.3\% to 41.1\% and 50.8\% to 61.7\%, respectively (see the 4th row and the 5th row), but it is still poor. (3) Obviously, the STR models can only do recognition, cannot generate SR images. (4) STISR methods are proposed to boost the recognition performance by generating SR images. These methods succeed in lifting the recognition accuracy, for example, TG, TPGSR and C3-STISR boost the accuracy of CRNN from 27.3\% to 48.9\%, 51.8\%, and 53.7\%, respectively (see from the 6th row to the 11th row). Nevertheless, STISR approaches also have the risk of misleading the recognition. For instance, as can be seen in the 5th and the 9th rows, the performance of TG+SVTR (61.6\%) is inferior to that of finetuning SVTR (61.7\%).

Then, we pay attention to the performance of our method IMAGE. To better demonstrate the performance of IMAGE under various computational requirements, two variants are designed, namely IMAGE-1 (the 12th and 13th rows) and IMAGE-2 (the 14th and 15th rows). From the 12th to the 15th rows of Tab.~\ref{tab:ic15-352}, we can see that (1) IMAGE-1 is good enough to outperform state-of-the-arts on recognition accuracy. In particular, IMAGE-1 has an accuracy of 68.6\%, which is 1.5\% higher than C3-STISR+SVTR (the 11th row). It is worth mentioning that our method is much better at recognizing hard samples. Specifically, the improvements of IMAGE-1 on \emph{Medium} and \emph{Hard} settings (5.1\% and 5.5\%) are higher than that on \emph{Easy} setting (0.2\%). This indicates the strong ability of IMAGE on reading low-resolution scene text images. (2) The best fidelity performance is also obtained by our technique. In particular, IMAGE-2 (the 15th row) lifts the PSNR from 21.51 to 21.92. (3) By taking the computational cost into consideration, the advantage of our method are better demonstrated. Consider that all the methods with FLOPs lower than 2.0G, IMAGE-1 achieves the best recognition accuracy and the second-best fidelity. Our IMAGE-2 obtains both the best recognition accuracy and the best fidelity, while decreases the FLOPs from 5.53G to 3.41G and Params from 72.86M to 56.22M. These results indicate the superiority of IMAGE on recognition accuracy, fidelity performance, and computational cost. We do not present results of IMAGE-3. The reasons are two-fold: (1) this version has a Params of 80.6M, which exceeds the Params of other methods; (2) IMAGE-1 and IMAGE-2 are already good enough to outperform the state-of-the-arts.

We also present the running time measured by an RTX3090 GPU to better demonstrate the advantages of our IMAGE method. Related results are given in Tab.~\ref{tab:speed}. Obviously, our IMAGE-1 runs faster than TPGSR and C3-STISR. Although IMAGE-1 is a bit slower than SVTR, it obtains 6.9\% higher accuracy and can output high fidelity SR images.
\begin{table}[t]
\centering
\scalebox{0.85}{
\begin{tabular}{c||c|c|c|c||c|c}
\hline
~ &CRNN&SVTR&TPGSR &C3-STISR &IMAGE-1&IMAGE-2\\
\hline
Speed (ms) &3.4&21.5&74.0&41.6&35.5&62.7\\
\hline
\end{tabular}}
\caption{Running speed (measured on a RTX3090 GPU) comparison. CRNN is used as the recognizer for STISR.}
\label{tab:speed}
\end{table}

\begin{figure*}
	\begin{center}
		\includegraphics[width=0.98\linewidth]{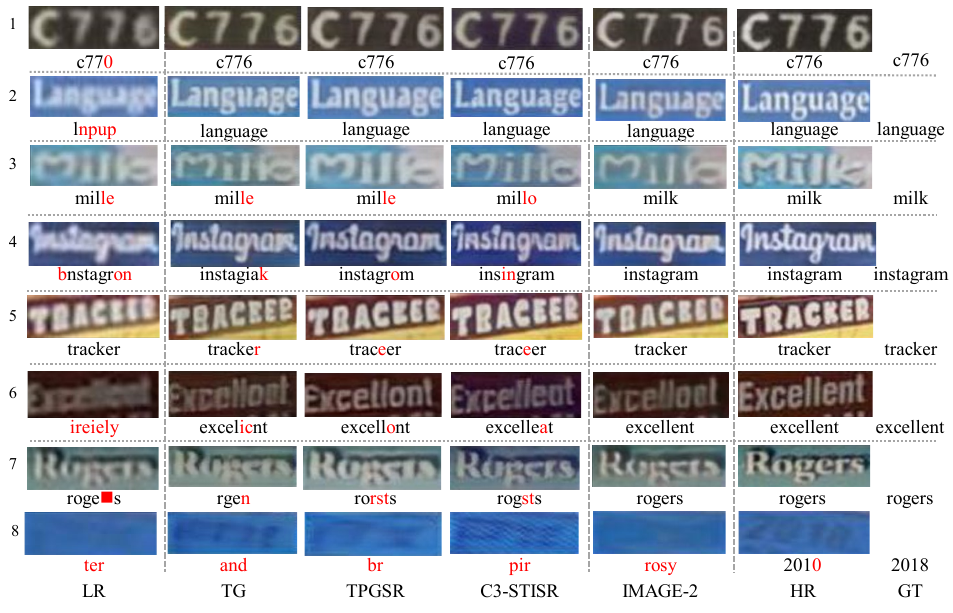}
	\end{center}
	\caption{Examples of generated images and recognition results. Here, GT is ground truth text. Red/black characters are incorrectly/correctly recognized. Texts below pictures in the LR, TG, TPGSR, C3-STISR and HR columns are recognized by SVTR.}
	\label{fig:visualization}
\end{figure*}
\begin{table}[t]
\centering
\scalebox{0.98}{
\begin{tabular}{c|c|c|c}
\hline
Method &Accuracy&FLOPs~(G) & Params~(M)\\
\hline
SVTR&66.5\%&1.90&22.70\\
\hline
TG &66.4\%&2.81&23.53\\
TPGSR &71.3\%&5.74&49.67\\
C3-STISR &75.0\%&5.53 &72.86\\
\hline
IMAGE-1 &\textbf{76.7}\%&1.98&31.84\\
IMAGE-2 &\textbf{80.7}\%&3.41&56.22\\
\hline
\end{tabular}}
\caption{Performance comparison with various state-of-the-art baselines on IC15-352 dataset. The recognizer used for STISR methods is SVTR.}
\label{tab:ic15-352}
\end{table}

\subsubsection{Performance improvement on IC15-352}
We also report the performance of typical approaches on the IC15-352 dataset. Since there are not LR-HR image pairs in this dataset, we only report accuracy to evaluate the recognition performance. The experimental results are given in Tab.~\ref{tab:ic15-352}. Some similar conclusions can also be drawn. For example, TG+SVTR also deteriorates the recognition performance. And obviously, the best recognition performance is achieved by our method with a low computational cost.

\subsection{Visualization}
We have conducted extensive quantitative experiments to show the superiority of IMAGE over existing approaches. Here we conduct a visualization study to better demonstrate the advantage of our method. 8 representative cases are given in Fig.~\ref{fig:visualization}. The LR and HR columns indicate the input LR and HR images, and the texts below them are directly recognized by SVTR. We can see that (1) typical recognizers cannot recognize well low-resolution text images (see the 1st and the 2nd cases) while STISR methods and our method can do recognition better. (2) The 3rd - 5th cases show one major weakness of typical STISR methods. Obviously, STISR approaches may generate erroneous results and mislead the subsequent recognition. Taking the 3rd case for instance, TG, TPGSR, and C3-STISR mistakenly recover the last character `k' to `e', `e', and `o', respectively. What is worse, for the 5th case, we would rather drop STISR and directly let the recognizer guess the blurry pixels. Fortunately, our method only implicitly uses the super-resolution results as clues to help recognition, which avoids the explicit propagation of errors. (3) As a result, as can be seen in the 6th and the 7th cases, IMAGE can still correctly read some tough cases where the information from the super-resolution model is insufficient or even erroneous (e.g. the recovery of `n' in the 6th case fails). (4) Another limitation of STISR approaches is that they may hurt the fidelity to make the SR images much more distinguishable. For example, in the 7th case, TPGSR slightly over-lightens the background while C3-STISR deepens the shadow of the text to highlight the text. (5) Finally, as can be seen in the 8th case, in some extremely tough cases where IMAGE fails to capture sufficient information from LR images and clues, the recognition and the recovery will fail.
\begin{table}[t]
\centering
\scalebox{0.98}{
\begin{tabular}{c|c|c|c}
\hline
\multirow{2}*{Method} & \multicolumn{3}{c}{Metrics}\\
\cline{2-4}
~&Accuracy&PSNR(dB)&FLOPs(G)\\
\hline
IMAGE-2&70.3\%&21.92&3.41\\
Recognition model&65.6\%&-&1.23\\
Super-resolution model&-&21.67&1.17\\
Without losses of IMAGE&57.5\%&21.33 &3.41 \\
\hline
\end{tabular}}
\caption{Ablation study on the proposed iterative mutual guidance.}
\label{tab:ab}
\end{table}
\begin{table}[t]
\centering
\scalebox{0.98}{
\begin{tabular}{c|cc|c|c}
\hline
\multirow{3}*{$i$} & \multicolumn{4}{c}{Metrics}\\
\cline{2-5}
~&\multicolumn{2}{c|}{Accuracy}&PSNR(dB)&FLOPs\\
\cline{2-4}
~&$p_i$&$\hat{p}_i$&$I_{i}^{SR}$&(G)\\
\hline
1 &57.5\%&57.1\%&21.13&1.70\\
2 &67.3\%&61.4\%&21.29&3.14\\
3 &70.3\%&-&21.92&3.41\\
\hline
\end{tabular}}
\caption{Performance of the $i$-th intermediate results of IMAGE-2.}
\label{tab:inter_results}
\end{table}
\begin{figure}
	\begin{center}
		\includegraphics[width=0.55\linewidth]{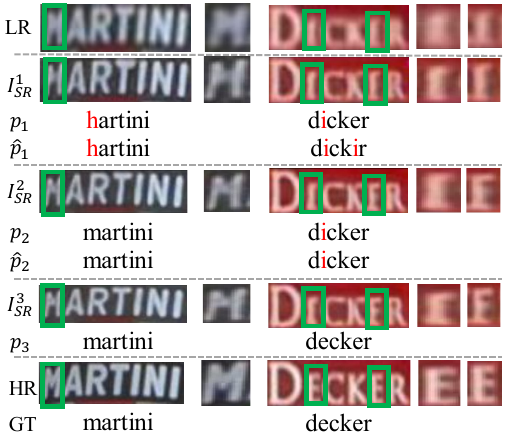}
	\end{center}
	\caption{Examples of the intermediate recognition results and SR images of IMAGE-2. Here, GT indicates ground truth. Red/black characters indicate
incorrectly/correctly recognized.}
	\label{fig:inter}
\end{figure}

\subsection{Ablation study}
Here we conduct an ablation study to check the advantage of IMAGE. We start by develop a variant without iterative mutual guidance, which means the recognition model and the super-resolution model are optimized individually without any interaction. As can be seen in the 2nd and 3rd rows of Tab.~\ref{tab:ab}, all the performances without IMAGE technique are deteriorated and IMAGE uses only an additional cost of 1.01G FLOPs to lift the accuracy from 65.6\% to 70.3\% and PSNR from 21.67 to 21.92. This shows the effectiveness of the proposed IMAGE. Besides, we also design a variant that removes the losses calculated for the supervision of the clues introduced in Sec.~\ref{sec:img}. We can see in the 4th row of Tab.~\ref{tab:ab}, the performance is significantly degraded. This validates the importance of our loss functions that can make clues much more informative.

Then, we report performance of the intermediate results of IMAGE-2 to better demonstrate the effectiveness of the clues generated by IMAGE. The metrics including the recognition performance of $\{p_{i}\}_{i=1}^{L+1}$ and $\{\hat{p}_{i}\}_{i=1}^L$, the fidelity of $\{I_{SR}^i\}_{i=1}^{L+1}$, and the cumulative computational cost. We can see in Tab.~\ref{tab:inter_results} that (1) the corresponding performance of the intermediate results is improved step by step with the help of the cross-modal clues. (2) Reading $\hat{p}_{i}$ that comes from the SR model is inferior to reading $p_{i}$ that is generated by semantic feature $h_{i-1}^s$ and clue $c_{i-1}^{p}$, which means that the SR images usually bring erroneous information and reading the texts with SR features as clue is a better solution. (3) It takes 0.26G FLOPs to offer initial features, 1.44G for each iteration, while 0.27G to decode the final results.

Finally, we also visualize the intermediate results in Fig.~\ref{fig:inter}. As can be seen in Fig.~\ref{fig:inter}, the intermediate results are improved step by step. In particular, in the left case, the character `h' is corrected in the second iteration while two `e's in the right case are also successfully recovered step by step. By taking these intermediate results as clues, the recognition and super-resolution are also lifted iteratively, which leads to more accurate final results.

\section{Limitation and future work}
The major limitation of IMAGE is that the generated SR images are targeted for fidelity. Thus, the SR images generated by IMAGE will be a bit less distinguishable, compared with typical STISR works (\textit{e.g.} the 7th case in Fig.~\ref{fig:visualization} and a bit lower SSIM). In addition, our method IMAGE has two separate models for STISR and STR, which makes the model a little complex, though it is not quite large. So as a future, we will try to pursue new solutions that do both recognition and recovery via a single model. 

\section{Conclusion}
In this paper, we propose a new method called IMAGE to do low-resolution scene text recognition and recovery in a framework. Different from existing STR and STISR works that directly read texts from LR images or recover blurry pixels to benefit recognition, IMAGE coordinates a text recognition model and a super-resolution model so that they can be optimized separately for their tasks while providing clues for cooperating with each other. IMAGE successfully overcomes the limitations of existing methods. Extensive experiments on two low-resolution scene text datasets validate the superiority of IMAGE over existing techniques on both recognition performance and super-resolution fidelity.




 \bibliographystyle{elsarticle-harv} 
 \bibliography{hire}






\end{document}